\pdfoutput=1

\documentclass[11pt]{article}

\usepackage{acl}

\usepackage{times}
\usepackage{latexsym}

\usepackage[T1]{fontenc}

\usepackage[utf8]{inputenc}

\usepackage{microtype}

\usepackage{paralist}

\usepackage{booktabs}
\usepackage{hyperref}
\usepackage{xcolor}		
 \definecolor{darkblue}{rgb}{0, 0, 0.5}
 \hypersetup{colorlinks=true,citecolor=darkblue, linkcolor=darkblue, urlcolor=darkblue}
\usepackage{url}
\usepackage{graphicx}
\usepackage{xcolor}

\definecolor{knolcol}{rgb}{0.0,0.2,0.4}
\definecolor{humancol}{rgb}{0.0,0.2,0.4}
\definecolor{robotcol}{rgb}{0.0,0.0,0.0}
\definecolor{topiccol}{rgb}{0.0,0.0,0.0}

\def\Snospace~{\S{}} 

\title{
Towards Responsible Natural Language Annotation \\for the Varieties of Arabic
}

\author{A. Stevie Bergman \\
  Responsible AI, Meta / New York, NY \\
  \texttt{asbergman@fb.com} \\\And
  Mona T. Diab \\
  Responsible AI, Meta / Seattle, WA \\
  \texttt{mdiab@fb.com} \\}
  
\begin{document}

\maketitle

\begin{abstract}
When building NLP models, there is a tendency 
to aim for broader coverage, often overlooking cultural and (socio)linguistic nuance. In this position paper, we make the case for care and attention to such nuances, particularly in dataset annotation, as well as the inclusion of cultural and linguistic expertise in the process. 
We present a playbook for responsible dataset creation for polyglossic, multidialectal languages. 
This work is 
informed by a study on Arabic annotation of social media content.

\end{abstract}

\section{Introduction}
\label{sec:intro}

Natural language processing (NLP) is the foundation of numerous automated decision-making systems in a growing number of scenarios and languages, including content moderation on platforms with global reach and consequence \citep{gillespie2020}. Thus it is  
highly pertinent to address how practitioners can build responsible NLP systems, models, and workflows for languages beyond English \citep{bender2019_benderrule, mielke2016, husain2021_survey}. 
In doing so, it is essential that these systems are designed with the inclusion of domain experts and stakeholder groups with native fluency and local/regional knowledge \citep{bender2009, ovadya2019}. 
This will not only ensure the presence of the deep problem understanding necessary to create accurate systems and anticipate potential harms, but foster earned trust and legitimacy in the system \citep{martin2020}.

Many performant machine learning/NLP algorithms to date are supervised, relying on large scale annotated training data, thus the veracity and curation of the data labels can have significant impact on model performance \citep{bender2021stochasticparrots, northcutt2021}. And further, even unsupervised and semi-supervised systems require labeled data for evaluation as a bare minimum to allow visibility into any blindspots in the system performance.

In this position paper, we focus on NLP datasets, highlighting the potential for compounding harms to at-risk populations and calling for greater care and attention to annotation and annotator support \citep{denton2021}. 
This is demonstrated via the varieties of Arabic.\footnote{We primarily use the term "varieties" of Arabic to refer to what are often colloquially called Arabic "dialects," "forms" of Arabic, or Arabic "languages." Referring to the varieties of Arabic as "dialects" does not do justice to (and even obfuscates) the fact that many are mutually incomprehensible (\S\ref{sec:arabic}). On the other hand, using the term "languages" minimizes the close connection between the Arabic varieties.} 

\section{Arabic Varieties and NLP}\label{sec:arabic}

Arabic is a Semitic language, spoken by over 420M people 
globally with the highest concentration in the Middle East and North Africa (MENA) where it is the dominant or official language of over twenty nations. 
Arabic is better described as a family of languages as the so-called "dialects" are highly variable and often have low inter-dialect comprehensibility \citep{arsocio2018}. \emph{e.g.} Moroccan and Egyptian varieties are about as mutually intelligible as Spanish and Romanian. 
Importantly, Arabic should not be considered as an analog to English, \textit{i.e.} one language with closely related dialects (\emph{e.g.} British, American, and Australian English). This comparison buries the polyglossic (when languages co-exist in a community) and heterogeneous characteristics of Arabic that are crucial to take into account when building effective Arabic NLP systems. We describe several relevant features of Arabic in this section.

First, while these varieties share the same root in Classical Arabic, varying historical and cultural experiences across the Arab world have led to the divergence in spoken practice, leading to frequent cases of "faux amis," \emph{a.k.a.} false cognates. This phenomenon refers to words or phrases that appear or sound the same between the varieties but have different meanings. For example, the word \textit{zahry}\footnote{We use the \href{https://en.wikipedia.org/wiki/Buckwalter_transliteration}{Buckwalter transliteration} standard to render Romanized Arabic text throughout the paper.} means \textit{my luck} in Tunisian Arabic, however the same word refers to the color 
pink in Levantine. 
Faux amis have implications for detecting objectionable speech, \emph{e.g.} a common slur in Moroccan Arabic is \emph{zamel}, however in Yemeni it is a type of singing that has become popular in recent years with the war in Yemen. 
Notably, this phenomenon is not only on the lexical level but also on the phrasal level, \emph{e.g.} \emph{yETyk AlEfyap} in Levantine dialects means "may you have good health," however in Moroccan it translates to "go to hell." 

Second, Arabic exemplifies one of the best known cases of diglossia,\footnote{\emph{Diglossia} refers to polyglossia with only two varieties.} 
where the formal Modern Standard Arabic (MSA) or \emph{fuSHY}, 
and the regional spoken vernaculars co-exist in virtually every speech community. 
Today, MSA is used in practice primarily for international news and legal contracts. It is generally the language of education, and thus only fully accessible to those with a certain level of literacy and training. In fact, MSA is a formal language akin to Shakespearean English and is not spoken colloquially. Despite that, MSA is considered a relatively high-resource language \citep{bender2019_benderrule} due to its status as the shared language in the Arab world, as such it is often the variety used in Arabic language technologies. Moreover, it is the prevalent Arabic variety of language-learning courses across the world. Notably, L2 language learners of MSA often have trouble communicating with native Arabic speakers due to considerable divergences between MSA and spoken varieties. Due to these factors, when NLP systems are 
built for MSA a large portion of Arabic speakers cannot benefit fully from the technology, 
creating an access disparity between those with and without a more advanced education.
\footnote{In the Arab world, basic literacy levels are 80\% for adults over 15 years and 75\% for females-only. (The World Bank: \href{https://data.worldbank.org/indicator/SE.ADT.LITR.ZS?locations=ZQ}{Literacy rate, adult total (\% of people ages 15 and above) - Middle East \& North Africa}, 1973-2020.)}

The third feature to highlight is the fact that the colloquial varieties of Arabic greatly differ from one another along a geographic and social continuum, however general consensus splits Arabic into six broad groupings. Following \citet{habash2010_arabicnlp} these are: \textbf{Iraqi} (Iraq), \textbf{Levantine} (Syria, Jordan, Lebanon, Palestine), \textbf{Maghrebi} (Morocco, Tunisia, Algeria, Libya,\footnote{Libyan Arabic is often in its own dialect grouping.} Western Sahara, Mauritania), 
 \textbf{Peninsular/Gulf} (Oman, Qatar, Saudi Arabia, Bahrain, United Arab Emirates, Kuwait\footnote{Kuwait is sometimes placed with the Iraqi dialect.}), \textbf{Yemeni} (Yemen), \textbf{Egyptian} (Egypt \& Sudan\footnote{This Egyptian \& Sudani grouping is standard, yet questionable due to linguistic differences and further fraught when considering the history of the region \citep{troutt2003}.}). 

These broad dialect groupings of course do not fully account for the heterogeneous political and cultural aspects of MENA or the Arabic speaking diaspora. This can be a pitfall in annotation given that contextual knowledge is often necessary for full semantic and pragmatic understanding. As an example, the Maghrebi dialect may be more appropriately broken down further, as Algerian is often unintelligible to Moroccan and Tunisian variety speakers due to the infusion of novel, contextually-based vocabulary that correspond to the country's particular political context. 
This is a common phenomenon across MENA, and of course for other languages. 
The unique country-level situations for Arabic dialects -- not to mention contexts and groupings that are sub-national or transcend country borders -- are important to take into account as NLP practitioners strive to produce systems that capture how humans naturally use language.

A final element to mention is the fact that there are no standard orthographies for Arabic dialectal text and the written variants exhibit pervasive code switching. This property is of course particularly important in the context of text annotation. Such a language profile pushes the boundaries of NLP, and perceptions away from Arabic as a monolith.\footnote{For a well-structured review of Arabic language detection, see \citet{husain2021_survey}. For an intro text on Arabic NLP, see \citet{habash2010_arabicnlp}, and for a fundamental introduction to Arabic sociolinguistics, see \citet{arsocio2018}.}

\section{Motivational Case Study: Arabic on Social Media}
\label{sec:case_study}

To highlight 
the necessity for a more nuanced treatment of polyglossic, multidialectal languages, we study annotation of Arabic content on the Facebook platform. 
The text-based samples in the study are in native Arabic script (\emph{i.e.} non-romanized), created by users in ten Arabic nations: Algeria, Morocco, Libya, Syria, Iraq, Lebanon, Egypt, Saudi Arabia, Sudan, and Yemen. 
We select these nations to cover all broad dialect groups in the region (\S\ref{sec:arabic}), with some inter-dialect comparisons, \emph{e.g.} Syria and Lebanon. Country-level groupings are employed in the study, rather than the broad dialects, to capture the contextual nature of content in each nation. 

Approximately 3000 pieces of content are sampled per country 
from a source dataset created for the purpose of hate speech classification evaluation. 
For content from each country, we compare data labeled by annotators residing in Morocco (AiM) to reviews by annotators with native fluency and expertise for each country in the study, \emph{a.k.a.} country experts (CE). For example, for Syrian content, annotation by AiM is compared to that by CE who are verified Syrian annotators. It is worth noting that the CE annotation consistency as measured by intra CE agreement -- inter-annotator agreement or IAA -- is quite high ranging from 64\% for Egyptian to 92\% for Moroccan Arabic.\footnote{Variations in CE agreement is outside the scope of this investigation. For this, we defer to future research.} The overlap sample sizes for the CE IAA measurement ranged in size from 91 unique items corresponding to 26\% of the total Saudi Arabic dataset to 975 unique items corresponding to 41\% of the total Iraqi dataset. 

To achieve this, the content is first sampled and sorted primarily by IP into per-country queues. CE then choose the dominant language variety and country of the sample. From there, the data is filtered to only Arabic content from the country. After filtering, the CE 
label the content (previously labeled by AiM) with one of two classes: "delete" (\emph{i.e.} the content is deemed hateful) or "ignore" (\emph{i.e.} the content is deemed benign) per a predefined set of content guidelines.

\begin{figure}[t]
 \centering
 \includegraphics[trim=0.2cm 0.5cm 0cm 0cm, clip, width=\linewidth]{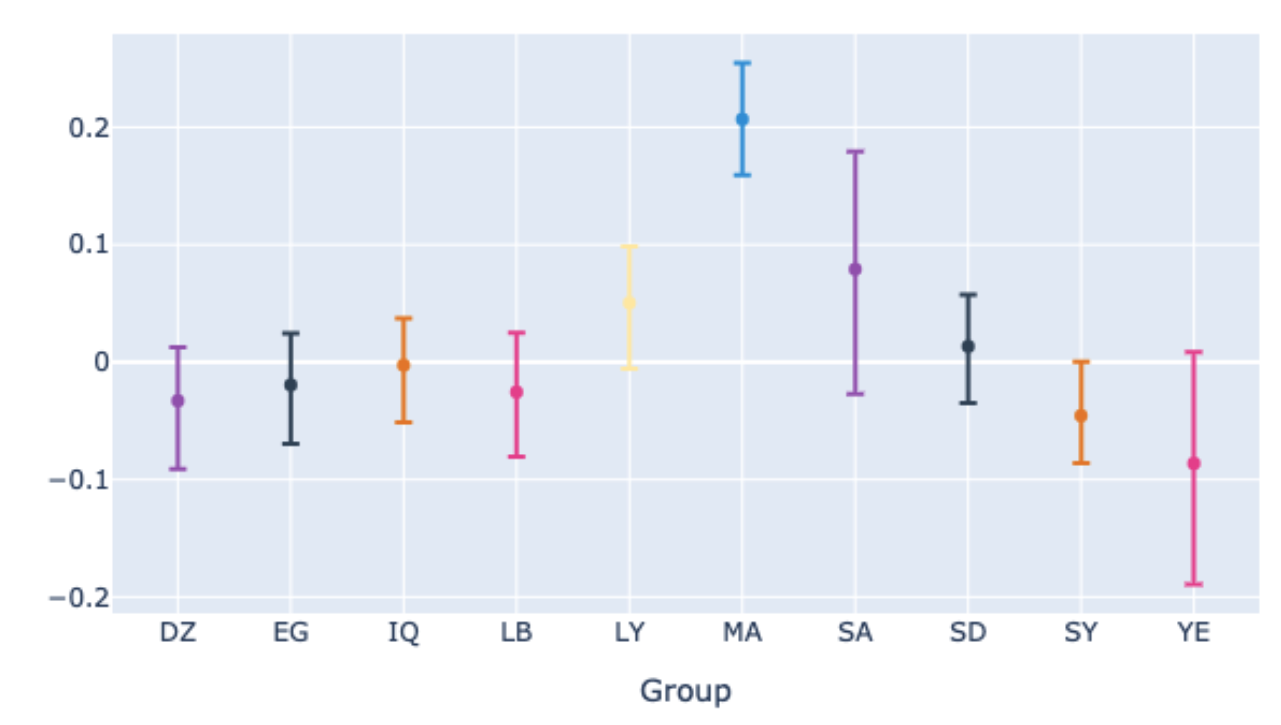}
  \caption{Aggregate implied threshold per country of the AiM using Signal Detection Theory (reduced by the overall aggregate implied threshold for all countries combined). 
  Note that a relatively \emph{higher} threshold indicates greater \emph{leniency}, or alternatively that a \emph{lower} relative threshold indicates greater \emph{strictness}.
}
  \label{fig:sdt}
\end{figure}

The main findings of this study are as follows: 
(1) For every country dataset, the majority of the content is found to be in the Arabic dialect of that nation, rather than MSA. This tracks with the understanding that users communicate informally on social media \citep{habash2010_arabicnlp, mcculloch2019_becauseinternet},
(2) The Saudi and Algerian datasets show a significantly higher presence of MSA content (31\% and 27\%, respectively) than other country datasets. Additionally, 74\% of the Arabic samples in the Saudi dataset are identified by CE as \emph{non-}Saudi content, a property of that could be explained by inward migration.\footnote{\href{https://data.worldbank.org/indicator/}{United Nations Population Division}, World Population Prospects: 2019 Revision.} Thus after filtering the Saudi dataset is significantly smaller than that of any other country dataset. 
(3) Signal Detection Theory (SDT) models decision-making as a mixture of Gaussians (one for benign content and another for violating content) and determines the \textit{implied} threshold of the reviewers in aggregate \citep{bakalar2020}. 
Employing SDT, we find the AiM reviewers are labeling Moroccan content more leniently with statistical significance (Figure~\ref{fig:sdt}) as compared to all other country datasets, with the notable exception of the Saudi dataset. 

Of further note, considering the CE labeling for each dataset as ground truth, the Morocco dataset had the most accurate labeling by the AiM cohort at 87\%. 
This is an intuitive result that was verified in this study. This finding, coupled with the SDT results shown in Figure~\ref{fig:sdt}, point to two potential interpretations: 
When reviewers understand the linguistic and contextual content of the samples, they 
\begin{compactenum}
    \item Are more likely to understand the nuances of the content that make 
    it benign (\emph{e.g.} sarcastic or idiomatic speech);
    \item May feel comfortable giving a certain benefit-of-the-doubt (\emph{a.k.a.} leniency) to the content creator, whereas they may not for groups in which they are not a member.
\end{compactenum}
This observed leniency, alone, \textit{could} imply that reviewers should not review content within their own variety, however when coupled with the sociolinguistic knowledge presented in Section~\ref{sec:arabic} on the distance between the varieties and the finding that, for the Morocco dataset, the the AiM cohort and CE had the highest agreement, these findings instead point to the fact that reviewers could be misunderstanding other-variety content or are relatively more strict on out-group member content. 

This study highlights the impact of Arabic variety differences on annotation. Accordingly we gather observations and formalize the process rendering it applicable across similar scenarios, namely, dataset creation for polyglossic, multidialectal languages such as Indo/Malay, varieties of German, etc. 

\section{A Playbook for Dataset Creation}
\label{sec:framework}

For accurate annotation of natural language content, it is important for NLP practitioners to consider the entire flow of content to the training or evaluation datasets.\footnote{As has \href{https://www.theverge.com/2020/11/13/21562596/facebook-ai-moderation}{previously been described} for some social media platforms, content can be added to manual (or human) review queues through proactive identification by an ML classifier or reporting from a user \citep{facebook2020_contentreview}. This content is then de-duplicated, ranked, and routed to the proper queue for manual inspection and labeling \citep{verge2020_contentreview}.} 
Here, we propose general recommended practices for annotation, with further information for 
Arabic NLP dataset creation, informed by the case study in Section~\ref{sec:case_study}. The guidelines aim to: 
(1) ensure there is sufficient expertise to -- at minimum -- understand the variety and context of the annotation samples; (2) route samples to experts who are best equipped to field and process them; and, (3) provide expert support and inclusion in the process of NLP design and implementation.
These recommendations could seem obvious to some, however they are worth crisply laying out  
to set expectations of expert inclusion and support in NLP development. They are as follows:

\noindent\textbf{(1) Data sample collection and curation} that is representative of the speech/orthographies of the user cohort for the developers' intended systems, and refreshed periodically to capture changing relevant events ensuring concurrent and temporal sensitivity \citep{devries2019_objectrecog, rancic2021_oversampling}. Such measures can reduce the potential for harms due to group under-representation in datasets \citep{fairmlbook, buolamwini2018gender}.

For the case study, we employed a stratified sampling technique to endeavor to capture the linguistic landscape of Arabic variety-speakers, and relevant context coverage for Arabic content from the countries and communities in focus.

\noindent\textbf{(2) Training materials and interface design} to support the annotators in understanding their task. For the case study (\S\ref{sec:case_study}), this included prototyping and iterations on the training materials and interface, with multi-stage feedback from a subset of annotators. Additionally, there was select translation of training materials and the annotation interface into the Arabic varieties to assist understanding (localization).\footnote{For example, in prototyping with experts, we changed the term "MSA" in annotation questions to the Arabic term "\emph{fuSHY}."}


\noindent\textbf{(3) Annotator representation:}  Reviewers are adequately representative of the users whose content they are labeling, and have necessary fluency in the language and context of the samples they are vetting. Ideally, proficiency is validated with testing and responsible hiring practices.\footnote{
If other languages are needed for labeling, \emph{e.g.} to understand the training materials, examples, or prompts, proficiency standards are recommended for those languages as well (if the instructions are in English while the actual text to be annotated is Swahili, for instance).}

\noindent\textbf{(4) Annotator proficiency assessment} to verify both language proficiency 
and an understanding of context and culture to the relevant level for the task at hand. This has the added benefit of demonstrating legitimacy and trustworthiness of the dataset. Evaluations should be designed carefully as many speakers of one variety have passive knowledge (vs. deep/native understanding) of other varieties. Arabic speakers may consider themselves fluent enough in the passive variety however it is important that their proficiency level is carefully assessed.\footnote{Arabic speakers often have passive knowledge of Egyptian Arabic, the dominant variety in Arabic language media. However passive learning from media can lead to missing Egyptian specific vernacular/cultural nuance.}

Furthermore, evaluations per-variety in many of these polyglossic multidialectal language families 
should include contextual elements 
relevant to the country or region. At present, many out-of-the-box Arabic proficiency tests are specific to MSA, which is not recommended unless employed as a supplement. 

For the case study (\S\ref{sec:case_study}), a major difference between the Morocco-based annotators and country experts is their verified relevant expertise. 
The country experts are proficient in the Arabic variety and regional contexts relevant to the labeling tasks, with sufficient cultural understanding to accurately interpret content from the community.
Broad groupings (\S\ref{sec:arabic}) \emph{could} be employed, depending on the granularity of contextual content and goals of the model. Setting aside the vast in-homogeneity of cultures and complex contexts across the Arabic-speaking world, there can be a misconception that if an individual can speak one Arabic dialect/variety they can accurately label content for any or adjacent varieties. This should not generally be assumed \citep{arsocio2018, habash2010_arabicnlp}. 

\noindent\textbf{(5) Sample routing and queues} are recommended such that annotators are reviewing content within their expertise. For systems aiming to cover multiple dialects or language groupings (\emph{e.g.} multiple Arabic varieties), content could be divided by a manual or automated language identification system and routed to designated annotator queues.

For Arabic varieties, queues and routing at least at the granularity level of each broad grouping (\S\ref{sec:arabic}) is generally essential, due to the low inter-variety comprehension. If country context is deemed relevant for the type of samples and application,  \emph{e.g.} political contexts/social value content, 
further subdivisions (\emph{e.g.} country, age, gender identity, etc.) are likely to be important for accuracy and consistency in labeling. 
For systems with automated routing, language and geographic identification could be employed to detect and separate dialects and contexts. 
A possible starting point is the groupings as described in \S\ref{sec:arabic}, enhanced with guiding per-variety word lists.

These elements bring up the difficult question of granularity: Not just how, but \textit{how deeply} should practitioners divide datasets and annotation, in order to ensure sufficient coverage and understanding of the content? This is a deceptively challenging question as languages most commonly exist on a continuum and practitioners can often divide the natural language groups further with no clear ideal stopping point. This question of granularity presents the need for a careful deliberation of trade-offs, often around resource expenditure vs. annotator expertise and data coverage. 
Data availability or the cost of obtaining high-quality labeling will frequently become a limiting factor. 
The over-arching recommendation in this work is to prioritize high quality annotation over breadth of coverage, indicating that in the cases where researchers do not have the resources for high quality annotation and annotator support they instead reduce coverage by tightening the goals of their NLP technology/application.

\noindent\textbf{(6) Sample re-routing capabilities} are relevant to capture routing errors, especially as errors in language identification can be surprising. It's important to build flexibility into the annotation system to allow 
annotators to route/skip samples outside their expertise, as well as a mechanism to surface such routing failures to improve the system. 
For the case study, we provided a mechanism to filter or surface routing errors.

\noindent\textbf{(7) Data evaluation} for faithful and accurately-labeled evaluation and training (if applicable), such that a resulting model can report real patterns \citep{northcutt2021}. This includes label evaluation and auditing systems to measure accuracy, sensitivity, and any potential bias (\emph{e.g.} sampling biases, statistical biases, or stereotypes) encoded in the datasets \citep{bakalar2020, barocas2016}. A multi-evaluation system for the stability of the labels, or uncovering ambiguity, could be significant \citep{caliskan17}.

For multidialectal, polyglossic languages such as Arabic, it is important that data evaluations account for the language  variety and contextual differences, particularly with tricky faux amis. For the case study, we employed a multistage process with adjudication using multiple expert reviewers of content to check annotation quality. 

\noindent\textbf{(8) Reviewer well-being support systems}, including but not limited to rest periods and psychological support, are important for any type of annotation, and especially-so when the dataset is composed of disturbing or traumatic content. 
These means of support are even more essential for the well-being of Arabic annotators, given the high rate of PTSD in the Arab world \citep{suto2016_trauma, syria2021_ptsd}.

\vspace{+1mm}
Care is needed when constructing Arabic natural language datasets, due to the prevalence of faux amis, Modern Standard Arabic and pervasive linguistic code switching, as well as the complex cultural and political contexts of Arabic-dominant countries. This is not to mention the potentially dire consequences of errors for economically disadvantaged and vulnerable groups in MENA \citep{amnesty2020_mena, hrw2017_gulfstates, tahrir2021_hate, brookings2021_yemen}. 
The guidelines in this section serve the additional purpose of surfacing failure modes in labeling that could scale when training classifiers, or obfuscate issues with model outcomes for evaluation datasets. 

\section{Conclusion}
\label{sec:conclusion}

Polyglossic and multidialectal languages present both challenges and opportunities to the NLP community, chief among them are the 
trade-offs inherent in dataset creation. 
From 
developing best practices with Arabic, we can apply the guidelines to other polyglossic, dialectal languages -- such as Chinese, Indonesian/Malay, and German -- with the understanding that with each language comes new challenges. 


\newpage

\section{Ethical Considerations}
\label{sec:ethical_considerations}
This paper 
fits in to the body of responsible and fair AI research by offering best practices towards ensuring the annotation ecosystem for dataset creation is responsive to the relevant groups in the Arabic-speaking world, and by extension local communities. 
The general goals of this work are to limit representational and potential downstream allocative harms \citep{fairmlbook}. Moreover, the work strives to advise on building NLP systems that allow annotators to be a voice for their local communities, where they are appreciated for their skill and ability to provide deep problem understanding.

\vspace{+.5mm}\noindent\textbf{Expert Inclusion:} Regional/country experts with Arabic variety proficiency are included in all stages of the guideline formulation, research design and implementation, and 
producing the comparative annotation results in Section~\ref{sec:case_study}. 
Their expertise is critical 
to the entire research process.\footnote{Due to safety concerns, we have not included these experts in authorship, and cannot include them all by name in acknowledgments. Instead, we have acknowledged them personally and professionally where possible.}

\vspace{+.5mm}\noindent \textbf{Audience:} This work is targeted for the NLP community. The guidelines are formulated with an understanding that practitioners may not have the resources to implement them all to the fullest extent. They are north stars to aim for, however if language/content understanding and reviewer support are not achievable with the resources at-hand, the practitioner can reduce the dialect/variety coverage of the annotation accordingly. We advocate for a nuanced approach in dataset creation over comprehensive coverage.

\vspace{+.5mm}\noindent\textbf{Scope:} This work is limited to Arabic varieties, with the hope that researchers can gain insight into handling other polyglossic, multidialectal global languages as well as a sense for the complexities of dataset creation for NLP. There are certainly nuances to Arabic annotation that are not covered in this work that affect annotation such as code switching and orthographic considerations. 

Furthermore, we focus on the broad Arabic dialects when discussing representation in the guidelines (\S\ref{sec:framework}). We recognize that there are other, deeply important areas of diversity and representation including gender, political stance, religion, etc., however we considered these outside the scope of this particular paper. What we describe here are recommended \emph{minimal} representation requirements considering language at as high of a level as possible. Of course other groupings have particular manners of language use that could be required for fully accurate annotation.

\vspace{+.5mm}\noindent\textbf{Broader Impacts:} NLP systems are embedded in our multifaceted, ever-changing societies, and it is therefore necessary to consider the model's potential or realized impacts, as well as the productive and adversarial manners in which the world can feedback to the model \citep{sambasivan2020, hagerty19_global}. The primary scope of this paper is data, however in what follows we discuss elements of model support that can provide constructive feedback to the system. 

First, this paper calls for soliciting stakeholder input, particularly through working with annotators who are of the community the model aims to capture. But further consultation with groups such as regional user-advocacy groups can be important to garner a higher-level view and broader problem understanding to prevent potential issues and low performance for underserved groups \citep{martin2020, caliskan17, bruckman2020haveyouthoughtabout, ovadya2019, abid21_muslim_bias}.

Other practices to aid the practitioner in envisioning the possible impacts of their work include: 
staged system roll-out or prototyping to get ahead of any unforeseen issues before full launch, and performing an impact investigation. Impact investigations are worthwhile, though they are neither simple nor straightforward and there are no clear norms \citep{prunkl2021_impact, pai_managingrisk2021}. 

The nature of statistical prediction means careful error handling is of the highest importance, as these systems will never be mistake-free, and in fact NLP systems can have surprising or unanticipated errors. In creating NLP systems, practitioners can ask what can be done to minimize potential negative impacts of errors 
\citep{hellman2019}. At the massive scales at which AI can operate, even a small error rate could affect many people \citep{sullivan2016}.

And, to garner constructive feedback, meaningful transparency measures are important \citep{diakopoulos2016accountability, mitchellmodelcards2019}, as are mechanisms for external feedback for model improvement in order to allow the model to be responsive to external events. 

These practices are generally important, but especially-so for the Arab world, as much of MENA is in-conflict, afflicted by ongoing tensions and political violence \citep{amnesty2020_mena, brookings2021_yemen} that can be amplified by technology \citep{tahrir2021_hate} or harnessed by violent and/or authoritarian state actors \citep{hrw2017_gulfstates}.


\section*{Acknowledgements}
This research would not have been possible without the invaluable expert guidance and contributions from colleagues from across the Arabic-speaking world, including Amine Dehimi, Hasan Ali, Hamoud Agha, Fajr Sabouni, Raed Jamil, Anthony Akoury, Abdelrahman ElSobki, Sarah Nasr, and Mohammad Zaher. The authors are sincerely grateful for their support.

The authors would like to further thank Amine Dehimi,  
Miranda Bogen, Molly FitzMorris, Khalid El-Arini, Edmund Tong, Renata Barreto, Jenny Hong, Jonathan Tannen, and Kate Vredenburg for the insightful conversations and feedback on the paper. Moreover, we would like to acknowledge valuable feedback from several anonymous ARR and ACL reviewers. 


\bibliography{anthology,custom}
\bibliographystyle{acl_natbib}

\newpage
\appendix 

\end{document}